\documentclass[runningheads]{llncs}

\usepackage{eccv}

\usepackage{eccvabbrv}

\usepackage{graphicx}
\usepackage{booktabs}

\usepackage[accsupp]{axessibility}  

\usepackage{hyperref}

\usepackage{orcidlink}

\usepackage{upgreek}

\newcommand{\textnum}[1]{$\mathrm{#1}$}
\newcommand{\0}{\textnum{0}}

\begin{document}

\title{Image color consistency in datasets: the Smooth-TPS3D method}


\author{Ismael Benito-Altamirano\inst{1,3}\orcidlink{0000-0002-2504-6123} \and
David Martínez-Carpena\inst{2}\orcidlink{0000-0001-9452-5315} \and
Hanna Lizarzaburu-Aguilar\inst{1,2} \and
Carles Ventura\inst{3} \and
Cristian Fàbrega\inst{1}\orcidlink{0000-0001-8337-4056} \and
Joan Daniel Prades\inst{4}\orcidlink{0000-0001-7055-5499}}

\authorrunning{I. Benito-Altamirano et al.}

\institute{
    MIND/IN2UB, Department of Electronic and Biomedical Engineering, Universitat de Barcelona, Carrer de Martí i Franquès, 1, Barcelona, 08028, Barcelona, Spain\\
    \and
    Department of Mathematics and Computer Science, Universitat de Barcelona, Gran Via de les Corts Catalanes, 585, Barcelona, 08007, Barcelona, Spain\\
    \and
    eHealth Center, Faculty of Computer Science, Multimedia and Telecommunications, Universitat Oberta de Catalunya, Rambla del Poblenou, 156, Barcelona, 08018, Barcelona, Spain\\
    \and
    Institute of Semiconductor Technology (IHT) and Laboratory for Emerging Nanometrology (LENA), Technische Universität Braunschweig, Hans-Sommer Str. 66, Braunschweig, D-38106, Germany\\
    \email{ismael.benito@ub.edu; ibenitoal@uoc.edu}}

\maketitle

\begin{abstract}
    Image color consistency is the key problem in digital imaging consistency when creating datasets. Here, we propose an improved 3D Thin-Plate Splines (TPS3D) color correction method to be used, in conjunction with color charts (i.e. Macbeth ColorChecker) or other machine-readable patterns, to achieve image consistency by post-processing. Also, we benchmark our method against its former implementation and the alternative methods reported to date with an augmented dataset based on the Gehler’s ColorChecker dataset. Benchmark includes how corrected images resemble the ground-truth images and how fast these implementations are. Results demonstrate that the TPS3D is the best candidate to achieve image consistency. Furthermore, our Smooth-TPS3D method shows equivalent results compared to the original method and reduced the 11-15\% of ill-conditioned scenarios which the previous method failed to less than 1\%. Moreover, we demonstrate that the Smooth-TPS method is 20\% faster than the original  method. Finally, we discuss how different methods offer different compromises between quality, correction accuracy and computational load.
\end{abstract}

\section{Introduction}

Color reproduction is one of the most studied problems in the audiovisual industry, that is present in our daily lives, long before today's smartphones, when color was introduced to the cinema, color analog cameras and color home TVs~\cite{Hunt2005}. In the past years, reproducing and measuring color has also become an important challenge for other industries such as health care, food manufacturing and environmental sensing. Regarding health care, dermatology is one of the main fields where color measurement is a strategic problem, from measuring skin-tones to avoid dataset bias~\cite{Newton2020} to medical image analysis to retrieve skin lesions~\cite{LI201866}. In other variety of fields like for environmental sensing~\cite{Fernandes2020}, colorimetric indicators are widely spread to act as humidity~\cite{Jung2016}, temperature~\cite{Seeboth2014} and gas sensors~\cite{Engel2021}.

Image consistency on captured datasets is a reduced problem of color reproduction. While \textit{color reproduction} aims at matching the color of a given object when reproduced in another device as an image (e.g. a painting, a printed photo, a digital photo on a screen, etc.), \textit{image consistency} is the problem of taking different images of the same object in different illumination conditions and with different capturing devices, to finally obtain the same apparent colors for that object. In this problem, the apparent colors of an object do not need to match its “real” spectral color,  they rather have to be just similar in each instance captured in different scenarios. In other words, all instances should match the first capture, or the reference capture, and not the real-life color~\cite{Shafer1985}. Therefore, image consistency is the actual problem to solve in the before-mentioned applications, in which it is more important to compare acquired images between them, so that consistent conclusions can be drawn with all instances, than comparing them to an actual reflectance spectrum.

\begin{figure}[h!t]
    \centering
    \includegraphics[width=0.65\linewidth]{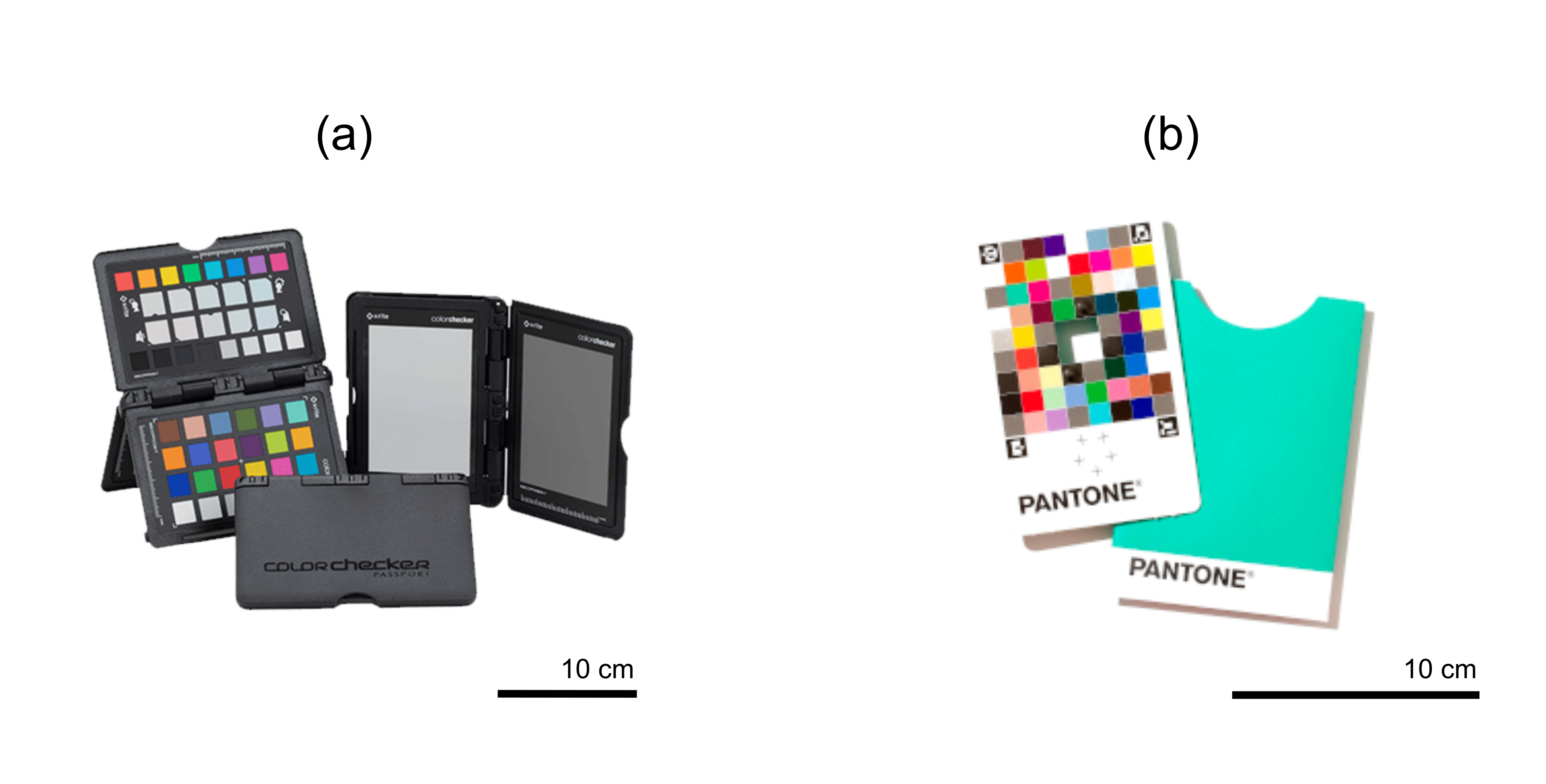}
    \caption{State-of-the-art color correction charts from X-Rite, Pantone. (a) The X-Rite ColorChecker Passport Photo 2\textregistered~kit. (b) The Pantone Color Match Card\textregistered.}
    \label{fig:colorcorrectioncharts}
\end{figure}

The traditional approach to achieve a general purpose color correction is the use of \textit{color rendition charts}, introduced by C.S. McCamy et. al. in 1976~\cite{McCamy1976} (see~\autoref{fig:colorcorrectioncharts}).

Color charts are machine-readable patterns placed in a scene that embed reference patches of a known color, where in order to solve the problem, several color references are placed in a scene to be captured and then used in a post-capture color correction process. The most simple color correction technique is the \textit{white balance}, that only involves one color reference. Beyond that, other techniques that use more than one color reference can be found elsewhere, using affine~\cite{Gong2013}, polynomial~\cite{Cheung2004, Finlayson2015}, root-polynomial~\cite{Finlayson2015} or thin-plate splines~\cite{Menesatti2012} transforms.



It is safe to say that, in most of these techniques, increasing the number and quality of the color references offers a systematic path towards better color calibration results. This strategy, however, comes along with more image area dedicated to accommodate these additional color references and therefore, a compromise must be found. This led X-Rite (a Pantone subsidiary company), to introduce improved versions of the ColorChecker, like the ColorChecker Passport Photo 2 ® kit. Also in this direction, Pantone presented in 2020 an improved color chart called Pantone Color Match Card®, based on the ArUco codes introduced by S. Garrido-Jurado et al. in 2015~\cite{Garrido-Jurado2016} to facilitate the location of a relatively large number of colors (see  ~\autoref{fig:colorcorrectioncharts}). And, in 2022, we introduced a new color chart based on QR Codes, which contains 24 color patches that can be read by any smartphone with a camera, and that can be used to color correct images directly in RGB spaces~\cite{BenitoAltamirano2023}. 

In this work, we focused on the implementation of thin-plate spline color corrections, specifically in the use of the TPS3D method to perform color correction in image datasets directly in RGB spaces, while proposing an alternative radial basis function~\cite{Buhmann2000} to be used to compute the TPS, and introducing a smoothing factor to approximate the solution in order to reduce corner-case errors~\cite{Rohr2001}, as we had experience using thin-plate splines for the readout of QR Codes in a 2D space~\cite{BenitoAltamirano2024}.  All in all, we illustrate here the advantages and limitations of the new TPS3D methodology to color correct RGB images.

Exploring the image consistency problem, requires an image dataset. The images on this dataset must contain some type of color chart, plus, the captured scenes in the images must be meaningful in day-life scenarios, and the color chart must be representative of those scenes. Instead of creating our own dataset, we proposed to use the widely-accepted Gehler’s ColorChecker dataset~\cite{Gehler2008, Hemrit2018}, which contains 569 images with a 24 patch Macbeth ColorChecker placed in different scenes.

\section{The Smooth-TPS3D method}

Color corrections are usually defined as mappings between color spaces, taking for example the definition of a color mapping between an RGB space~\cite{Gong2013}, namely $ (r, g, b) $ and another one $ (r', g', b') $:

 \begin{equation}
     r' = f_r (r, g, b)  = t_r + \sum_k^{r,g,b} a_{r, k} k
     \label{eq:affine}
 \end{equation}

\noindent where $ t_r $ is the translation contribution, $ a_{r,k} $ are the affine contributions, $f_r$ is the function which defines the relation between the red component of the $ (r', g', b') $ space and the three $ (r, g, b) $ color space components. These definitions can be expanded with non-linear terms, for example polynomial and root-polynomial terms, and can be found elsewhere~\cite{Cheung2004, Finlayson2015}.

\newpage
Menesatti et al. proposed using thin-plate splines, the TPS3D method, as the basis of the expansion to the $(r, g, b)$ color space~\cite{Menesatti2012}:

\begin{equation}
    r' = f_r (r, g, b)  = t_r + \sum_k^{r,g,b} a_{r, k} k + \sum_i^N w_{r,i} h_i (r, g, b)
    \label{eq:thinplate}
\end{equation}

\noindent
where $ t_r $ is the translation contribution, $ a_{r,k} $ are the affine contributions, $ w_{r,i} $ are the weight contributions for each spline contributions, and $ h (r, g, b) $ are radial-basis functions (RBF), $ h_i (r, g, b) $ are the kernels of $ h $ in the $ N $ known colors.

These RBF used to compute splines remain open to multiple definitions~\cite{Buhmann2000}. The thin-plate approach to compute those splines implies using solutions of the biharmonic equation:

\begin{equation}
    \label{eq:biharmonic}
    \Delta^{2} U = 0
\end{equation}

\noindent
where $U$ is the function that minimizes the bending energy functional described by many authors~\cite{whitbeck06,bookstein-89}, thus resembling the spline solution to the trajectory followed by an n-dimensional elastic plate.

\subsection{Polynomial radial basis functions}

The solutions for \autoref{eq:biharmonic} are the \emph{polynomial radial basis functions} and a general solution is provided for n-dimensional data as~\cite{Buhmann2000,bookstein-89}:

\begin{equation}
\begin{split}
   h_i(\mathbf{s}) & = U(\mathbf{s}, \mathbf{c}_i) =
   \begin{cases}
       || \mathbf{s} - \mathbf{c}_i ||^{2k-n} \ln|| \mathbf{s} - \mathbf{c}_i || & 2k-n \text{ is even} \\
       || \mathbf{s} - \mathbf{c}_i ||^{2k-n}  & \text{otherwise}
   \end{cases}
\end{split}
\end{equation}

\noindent
where $ n $ is the number of dimensions, $ k $ is the order of the functional, $ \mathbf{s} $ and $  \mathbf{c}_i $ are the color vectors --i.e. $ \mathbf{s} = (r, g, b) $-- between the splines are computed and $ || \cdot || $ is a metric. For a bending energy functional (the metal thin-plate approach) $ k = 2 $ and $ n = 2 $ (2D data), we obtain the usual thin-plate spline RBF:

\begin{equation}
     h_i(\mathbf{s}) = || \mathbf{s} - \mathbf{c_i} ||^2 \ln|| \mathbf{s} - \mathbf{c}_i ||
     \label{eq:rbf2D}
\end{equation}

But for $ k = 2 $ and $ n = 3 $ (3D data) we obtain:

\begin{equation}
    h_i(\mathbf{s}) = || \mathbf{s} - \mathbf{c}_i || 
    \label{eq:rbf3D}
\end{equation}

It is unclear why, in the TPS3D method to color correct images, Menesatti et al.~\cite{Menesatti2012} used the definition for 2D data (\autoref{eq:rbf2D}), rather than the actual 3D definition (\autoref{eq:rbf3D}) which according to the literature should yield to more accurate results.

For our method, the Smooth-TPS3D, we proposed using as a basis the 3D solution rather than the 2D one, we also present here the results on the impact of this change in the formal definition of the TPS3D.

So far, we have not defined a metric $ || \cdot || $ to solve the TPS contributions. We followed Menesatti et al.~\cite{Menesatti2012} and used the Euclidean metric of the RGB space. We will also name this metric $ \Delta_{RGB} $, as it is commonly known in colorimetry literature:

\begin{equation}
    || \mathbf{s} - \mathbf{c} || = \Delta_{RGB}(\mathbf{s}, \mathbf{c}) = \sqrt{(r_s -  r_c)^2 + (g_s -  g_c)^2 + (b_s - b_c)^2}
    \label{eq:rgb_distance}
\end{equation}


\subsection{Smoothing the thin-plate spline correction}

Approximating the TPS corrections is a well-known technique~\cite{Rohr2001}. Specifically, this is performed in ill-conditioned scenarios where data is noisy or saturated, and strict interpolation between data points, leads to important error artifacts. We propose now adding  a smoothing factor to the TPS3D, to improve color correction in ill-conditioned situations.

We approximated the TPS by adding a smoothing factor to the spline contributions, which reduces the spline contributions in favor of the affine ones. Taking \autoref{eq:thinplate}, we will introduce a smooth factor only for those color references where the center of the spline was those references themselves:

\begin{equation}
\begin{split}
    r'_j = t_r + \sum_k^{r_j,g_j,b_j} a_{r, k} k + \sum_i^N  w_{r,i} (h_i (r_j, g_j, b_j) + \lambda \delta_{ij})
\end{split}
\end{equation}

\noindent
where $ \lambda $ is the smoothing factor, and $ \delta_{ij} $ is a Kronecker delta. 

\section{Experimental pipeline}
\label{sc:ch6_experimental_details}

\subsection{Original dataset}

The Gehler’s dataset comprises images from two cameras: a Canon EOS 1DS (86 images) and a Canon EOS 5D (483 images), both cameras producing raw images of 12-bit per channel  with a RGGB Bayer pattern. This means we have twice as many green pixels than red or blue pixels. 


Images have been processed using Python and \texttt{rawpy}, the Python wrapper of \texttt{craw} binary, the utility used elsewhere to process the Gehler’s dataset~\cite{Gehler2008,Hemrit2018}. When developing the images, we implemented no interpolation, thus rendering images half the size of the raw image. These are our \textit{ground-truth} images: the colors in these images are what we are trying to recover when performing the color corrections.

We chose to work with 8-bits per channel RGB images as is the most commonly developed pixel format present nowadays. First, we cast the developed dataset 12-bit images to 8-bit resolution. The difference between the cast images and the ground-truth images is the \emph{quantization error}, due to the loss of color depth resolution.

\begin{figure}[h!t]
    \centering
    \setlength{\belowcaptionskip}{-10pt}
    \includegraphics[trim={0 5.2em 0 0},clip,width=0.5\textwidth]{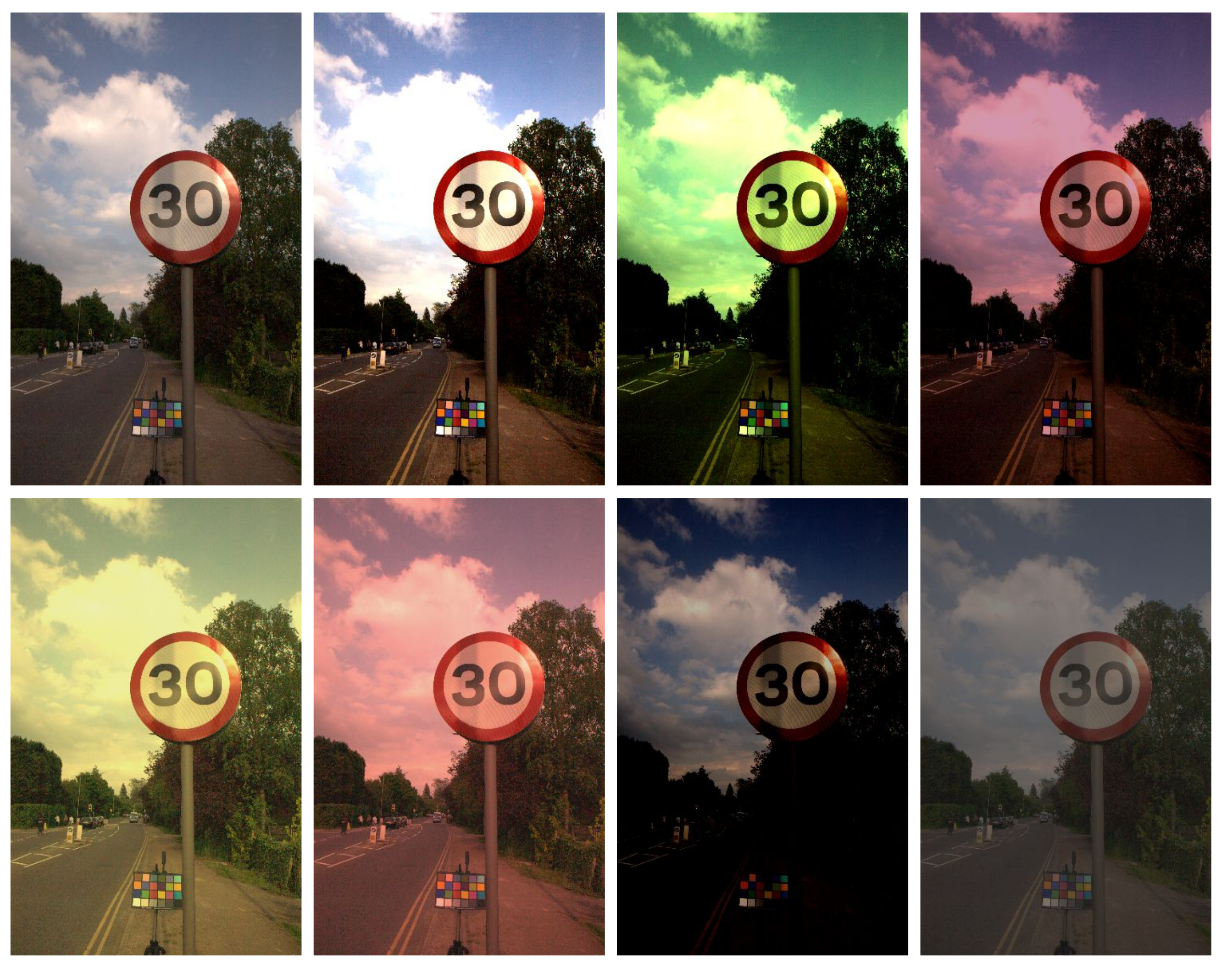}
    \caption{Different examples of color augmentation using \texttt{imgaug} in Python. The first image on the left is the developed original image from the Gehlre’s dataset. The other images are augmentations of these image with variations in color, contrast and saturation.  }
    \label{fig:ccdatasetaugmentation}
\end{figure}

\subsection{Image augmentation}

Subsequently, we augmented the dataset using Python and \texttt{imgaug}~\cite{imgaug} that generated image replicas simulating different acquisition setup conditions. The augmentations were performed with random augmentations that modeled: linear contrast, gamma contrast and channel cross-talk. In this augmentation process, we applied the same transformations to the whole image and to the ColorChecker. Geometrical distortions were omitted because this work is focused on a colorimetry problem (see~\autoref{fig:ccdatasetaugmentation}).

\subsection{Color corrections}



\noindent\textbf{A. Baselines:} First, a positive baseline (PERF), which represents the best correction we can have, which only source of error is the quantization error, from the developed image to the ground-truth image, which is expressed as the difference from 12-bit to 8-bit resolution. Then, a negative baseline (NONE), which represents not performing a correction at all. Represents the maximum error we can have in the color correction.

\noindent\textbf{B. Affine (AFF):} white-balance (AFF0), white-balance with black subtraction (AFF1), affine (AFF2), affine with translation (AFF3) from Gong et al.~\cite{Gong2013}. Contributions of these corrections were linear with the RGB values of the original image; i.e. for AFF3 the contributions are $ (1, r, g, b) $.

\noindent\textbf{C. Vandermonde (VAN):} four polynomial corrections from degree 2 to 5 (VAN0, VAN1, VAN2 and VAN3).  Contributions of these corrections are polynomial with the RGB values of the original image; i.e. for VAN2 the contributions are $ (1, r, g, b, r ^2, g ^2, b ^2) $.

\noindent\textbf{D. Cheung (CHE):} from Cheung et al.~\cite{Cheung2004}, four polynomial corrections with different terms: 5 (CHE0), 7 (CHE1), 8 (CHE2) and 10 (CHE3). Contributions of these corrections were polynomial, but also incluse cross-terms; i.e. for CHE2 the contributions are $ (1, r, g, b, rg, rb, gb, rgb) $.

\noindent\textbf{E. Finlayson (FIN):} from Finlayson et al.~\cite{Finlayson2015}, two polynomial and two root-polynomial, of degrees 2 and 3 (FIN0, FIN1, FIN2, FIN3). Contributions of these corrections were polynomial (FIN0, FIN1) and root-polynomial (FIN2, FIN3); i.e. for FIN2 the contributions are $ (r, g, b,  \sqrt{rg}, \sqrt{rb},  \sqrt{gb}) $.

\noindent\textbf{F. Thin-plate splines (TPS):}  TPS3D from Menesatti et al.~\cite{Menesatti2012} (TPS0), plus our method using the proper RBF (TPS1) and the same method with two smoothing values (TPS2 and TPS3), the full Smooth-TPS method. For TPS3 the contributions were $ (1, r, g, b,  \Delta_1,  \dots,  \Delta_{24}) $; where $ \Delta_i =  \Delta_{RGB} (\mathbf{s}_i, \mathbf{c}) $ are the spline contributions.



\subsection{Benchmark metrics}

\textbf{A. Within-distance $ \overline{\Delta_{RGB}}_{,within} $} as the mean distance of all and only the colors in the ColorChecker to their expected corrected values~\cite{Menesatti2012}:
\begin{equation}
  \overline{\Delta_{RGB}}_{,within} = \frac{\sum_{i=1}^{N} \Delta_{RGB} (\mathbf{s'}_i, \mathbf{c'}_i)  }{N}  
  \label{eq:within_distance}
\end{equation}
where $ \mathbf{s'}_i $ is the corrected version of a certain ColorChecker captured color $ \mathbf{s}_i $, which has a ground-truth reference value of $ \mathbf{c'}_i $, and $ N $ is the number of reference colors in the ColorChecker (in our case $ N = 24 $). Alongside with this metric, a criterion was defined to detect \textit{failed corrections}. We considered failed corrections those which failed to \textit{reduce the within-distance} between the colors of the ColorChecker after the correction. Then, by comparing the $ \overline{\Delta_{RGB}}_{,within} $ of the corrected image and the image without correction (NONE):
\begin{equation}
\overline{\Delta_{RGB}}_{,within} - \overline{\Delta_{RGB}}_{,within,NONE} > 0 \ .
\label{eq:within_comparisom}
\end{equation}


\noindent\textbf{B. Pairwise-distance set $ \mathbf{\Delta_{RGB}}_{,pairwise} $} as the set of the distances between all the colors in a ColorChecker in the same image:
\begin{equation}
    \mathbf{\Delta_{RGB}}_{,pairwise} = \left\{  \Delta_{RGB} (\mathbf{c'}_i, \mathbf{c'}_j ) \ : \ i,\,j = 1, \dotsc, N   \right\}
    \label{eq:pairwise_distance}
\end{equation}
where $ \mathbf{c'}_i $  and $ \mathbf{c'}_j $ are colors of the ColorChecker in a given image.  Also, we implemented another criterion to detect \textit{ill-conditioned corrections}. Ill-conditioned corrections are those failed corrections in which colors have also collapsed into extreme RGB values. By using the \textit{minimum pairwise-distance} for a given color corrected image:
\begin{equation}
    \min \left( \mathbf{\Delta_{RGB}}_{,pairwise} \right)  < \updelta \ ,
    \label{eq:pairwise_comparison}
\end{equation}
where $ \updelta $ is a constant threshold which tends to zero. Note that somehow we were measuring here the opposite to the first criterion: we expected erroneous corrected colors to be pushed away from the original colors \autoref{eq:within_comparisom}. However, sometimes they also got shrunk into the borders of the RGB cube \autoref{eq:pairwise_comparison}, causing two or more colors to saturate into the same color. Also, notice that we did not define a \textit{mean pairwise-distance}, $\overline{\Delta_{RGB}}_{,pairwise}$, as it was useless to define a criterion around a variable which presented huge dispersion in ill-conditioned scenarios (e.g. colors pairs were at the same time close and far, grouped by clusters). 



%

\noindent\textbf{C. Inter-distance $ \overline{\Delta_{RGB}}_{,inter} $} as the color distance between all the other colors in the corrected images with respect to their values in the ground-truth images (measured as the mean RGB distance of all the colors in the image but subtracting first the ColorChecker area as proposed by \cite{Hemrit2018}):
\begin{equation}
  \overline{\Delta_{RGB}}_{,inter} = \frac{\sum_{i=1}^{M} \Delta_{RGB} (\mathbf{s'}_i, \mathbf{c'}_i)  }{M}  
  \label{eq:inter_distance}
\end{equation}
where $ M $ is the total amount of pixels in the image other than those of the ColorChecker. This definition particularized the proposal of~\cite{Menesatti2012}, where in order to compute $ \overline{\Delta_{RGB}}_{,inter} $, they used  all the colors of another color chart instead of the actual image.


\noindent\textbf{D. Execution time $ \mathcal{T} $} to compare the computational performance of the methods, we measured the execution time to compute each corrected image, $ \mathcal{T} $ was also measured for images with different sizes to study its scaling with the amount of pixels in an image in all corrections~\cite{Luo2014}.

\section{Results}

\subsection{Detecting failed corrections}

Let us start with the results of the detection of failed corrections for each color correction  proposed. Here we used the defined criteria for $ \overline{\Delta_{RGB}}_{,within} $ (\autoref{eq:within_comparisom})  and $ \mathbf{\Delta_{RGB}}_{,pairwise} $ (\autoref{eq:pairwise_comparison}) to discover failed and ill-conditioned corrections.

First, we subtracted the $ \overline{\Delta_{RGB}}_{,within} $ measures to the other  $ \overline{\Delta_{RGB}}_{,within} $ and compared this quantity with \0, following \autoref{eq:within_comparisom}. Those cases where this criterion was greater than \0 were counted as failed corrections. Second, for those corrections marked as failed, the $ \mathbf{\Delta_{RGB}}_{,pairwise} $ criteria (\autoref{eq:pairwise_comparison}) was applied to discovery ill-correction scenarios in between failed corrections. The $ \mathbf{\Delta_{RGB}}_{,pairwise} $ criteria were implemented using a $ \updelta = \sqrt{3} $, due to the fact this is the $ {\Delta_{RGB}}_{,pairwise} $ of two colors that dist one digit from each other in each channel (i.e. (0, 0, 0) and (1, 1, 1) for colors in a 8-bit color space). Finally, we also computed the relative \% of failed color corrections referenced to the total of color corrections performed. This figure is relevant as we removed these cases from further analysis.

\begin{figure*}[!ht]
    \setlength{\belowcaptionskip}{-10pt}
    \centering
    \includegraphics[trim={0 0.25em 0 0},clip,width=\textwidth]{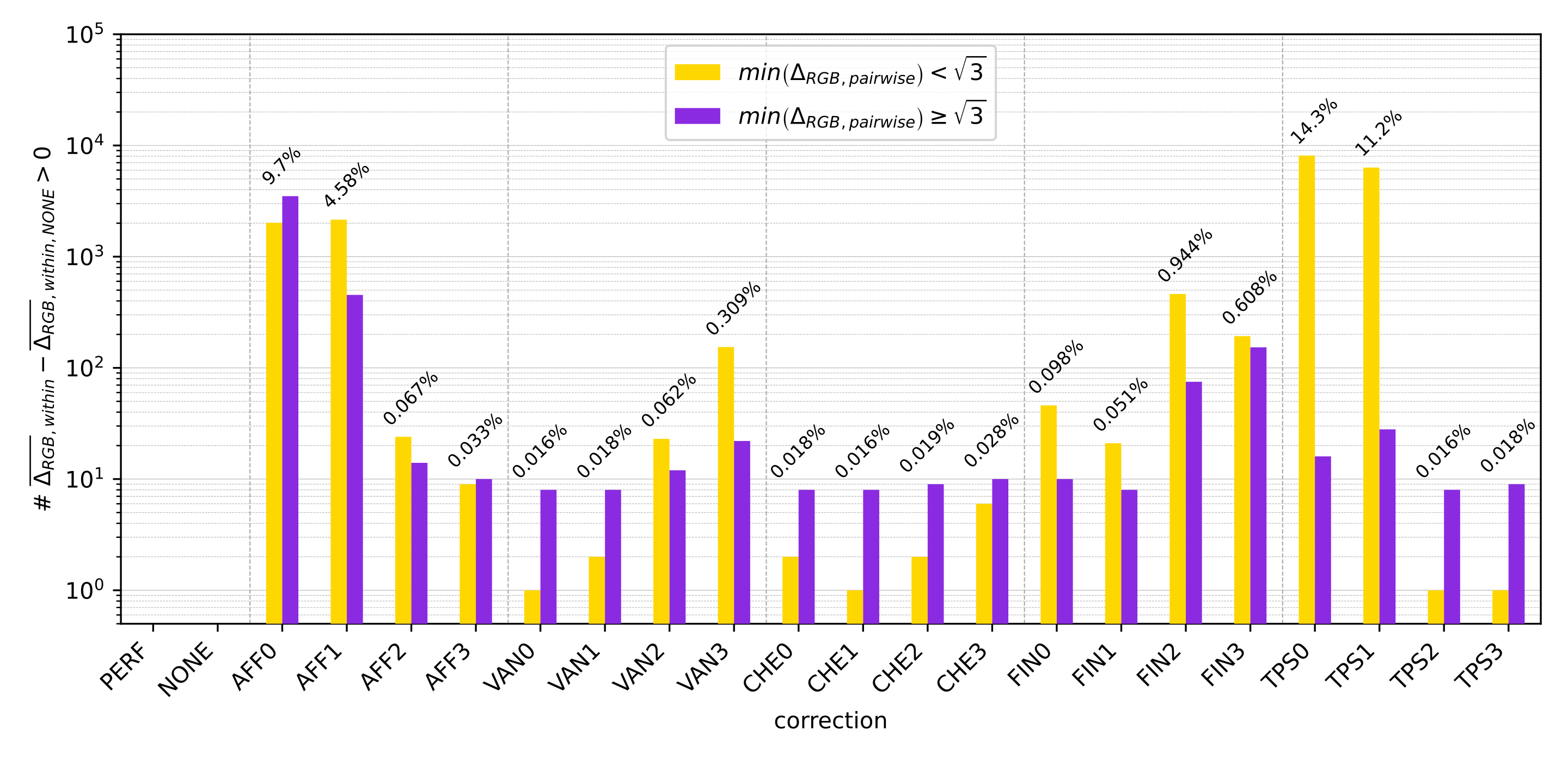}
    \caption{A count of the failed corrections for each correction method is shown. Failed corrections are selected if their $ \overline{\Delta_{RGB}}_{,within} $ computation is greater than the NONE correction. After this, the count is divided in ill-conditioned results or not. Ill-condition is assessed using the $ \mathbf{\Delta_{RGB}}_{,pairwise} $ comparison to a minimum distance of $\Delta_{RGB} = \sqrt{3}$.}
    \label{fig:corner_cases_count}
\end{figure*}

\autoref{fig:corner_cases_count} shows the results of this analysis. Results indicated that AFF transformations showed variations: AFF0 and AFF1 had high failure rates (9.7\% and 4.58\% respectively), while AFF2 and AFF3 had almost no failures due to using all available references. Polynomial methods (VAN, CHE and FIN) performed similarly, with all corrections having less than 1 \% failure and a correlation between the degree of polynomial expansion and failed corrections. TPS corrections had mixed outcomes, with TPS0 and TPS1 showing high failure rates (11.2 \% and 14.3 \% respectively), while TPS2 and TPS3 performed really well (less than 0.2 \% failure)  --our TPS implementations--. This suggests that our smoothing of the thin-plate contributions to the color correction was successful in fixing ill-conditioned scenarios.

\subsection{Color correction quality}

\begin{figure*}[!ht]
    \centering
    \setlength{\belowcaptionskip}{-10pt}
    \includegraphics[width=\textwidth]{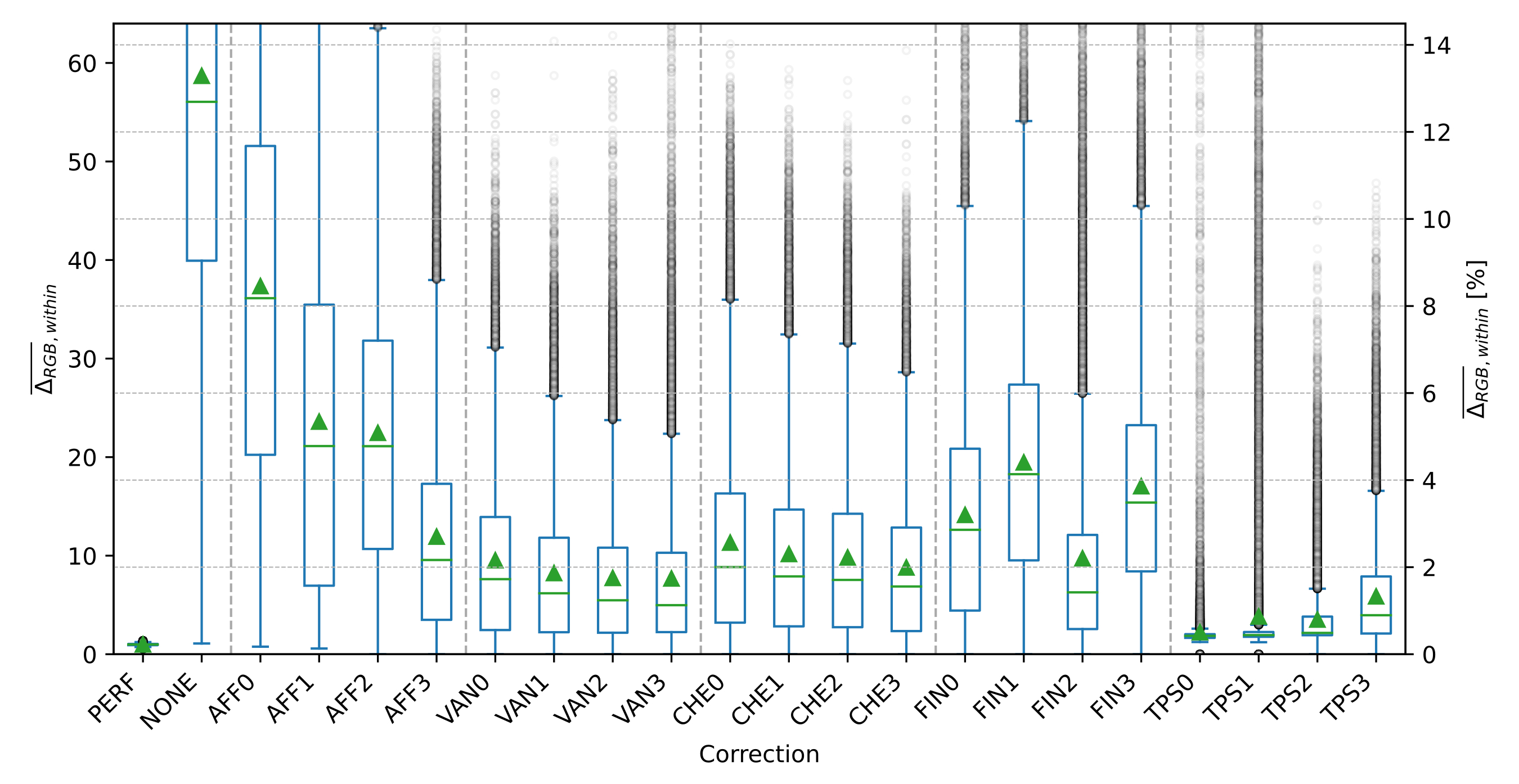}
    \caption{The $ \overline{\Delta_{RGB}}_{,within} $ for each image in the dataset is shown as a distribution against the color correction techniques. The distribution information is represented as follows: ($\triangle$) the mean, ($-$) the median at \textnum{Q2}, (box) from \textnum{Q1} to \textnum{Q3}, (whiskers) \textnum{1.5 \cdot (Q3 - Q1)} into \textnum{Q0} and \textnum{Q4} and ($\circ$) the outliers. PERF correction is not zero and shows the quantization effect. NONE is a reference of not applying any correction at all. The rest of the corrections are grouped in: AFF, VAN, CHE, FIN and TPS corrections.}
    \label{fig:within_distances_zoom}
\end{figure*}

\begin{figure*}[!ht]
    \centering
    \setlength{\belowcaptionskip}{-10pt}
    \includegraphics[trim={0 0em 0 0 },clip,width=\textwidth]{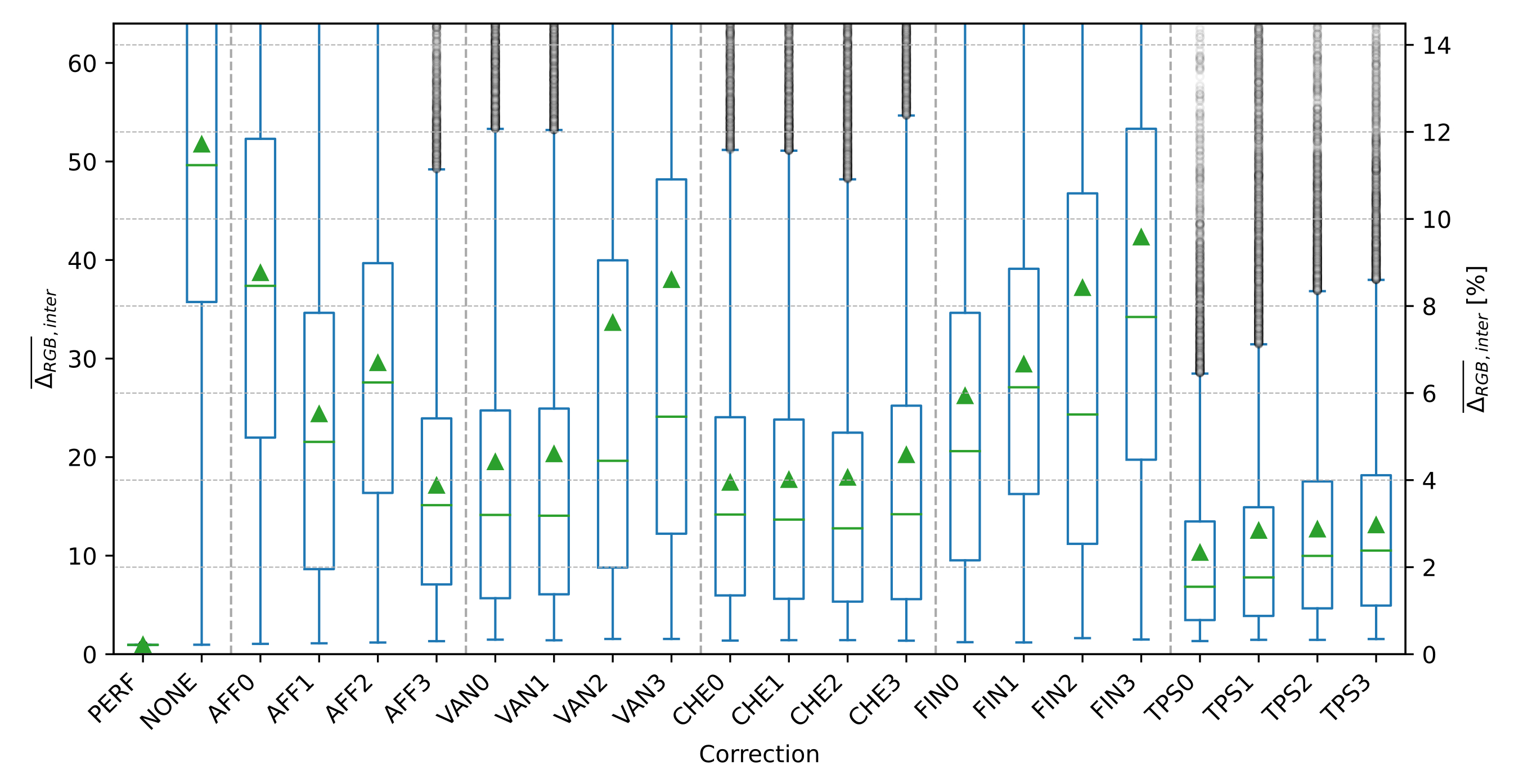}
    \caption{The $ \overline{\Delta_{RGB}}_{,inter} $ for each image in the dataset is shown a distribution against the color correction techniques. The distribution information is represented as follows: ($\triangle$) the mean, ($-$) the median at \textnum{Q2}, (box) from \textnum{Q1} to \textnum{Q3}, (whiskers) \textnum{1.5 \cdot (Q3 - Q1)} into \textnum{Q0} and \textnum{Q4} and ($\circ$) the outliers. PERF correction is not zero and shows the quantization effect. NONE is a reference of not applying any correction at all. The rest of the corrections are grouped in: AFF, VAN, CHE, FIN and TPS corrections.}
    \label{fig:inter_distances_zoom}
\end{figure*}

Once evaluated and cleaned the failed corrections from our results, we proceeded to evaluate how the proposed color corrections scored in terms of color correction quality. In other words, we evaluated how they minimize the median value of the \textit{within-distances distributions} and the \textit{inter-distances distributions}. Before, we defined $\overline{\Delta_{RGB}}_{,within}$ and $\overline{\Delta_{RGB}}_{,inter}$ similar to Menesatti et al.~\cite{Menesatti2012}, but it is clearer to understand these metrics with a percentage definition. So, we can normalize them by using the  maximum distance in the RGB space is set by the distance between the black and white colors ($255 \cdot \sqrt{3}$):

\begin{equation}
    \Delta_{RGB} [\%] = 100 \cdot \frac{\Delta_{RGB}}{\Delta_{RGB_{max}}} = 100 \cdot \frac{\Delta_{RGB}}{255 \cdot \sqrt{3}}
    \label{eq:rgb_distance_rel}
\end{equation}

both results in \autoref{fig:within_distances_zoom} and \autoref{fig:inter_distances_zoom} present the results with and without this normalization.



On one hand, results for \emph{within-distances} $\overline{\Delta_{RGB}}_{,within}$ (\autoref{fig:within_distances_zoom}) revealed that AFF corrections performed variably. AFF0, relying solely on white balance, had the worst performance with a mean $\overline{\Delta_{RGB}}_{,within}$ exceeding 8\%. AFF1 and AFF2 performed similarly with means above 5\%, while AFF3, the most complete affine correction, achieved around 3\%, demonstrating that adding a translation component improves results.

For polynomial corrections, VAN corrections consistently outperformed AFF, with all four scoring around 2\% or less. Increasing the degree of polynomial expansion in VAN corrections resulted in better fitting, converging to a minimum median $\overline{\Delta_{RGB}}_{,within}$ around 1\%. CHE corrections also outperformed AFF3 but were slightly worse than VAN, scoring between 1\% and 3\%. The addition of cross-term contributions in CHE did not significantly improve RGB color space deformation fitting. Surprisingly, FIN corrections performed the worst among polynomial corrections, with scores between 3\% and 5\%, likely due to the absence of translation components. Root-polynomial corrections (FIN1 and FIN3) scored approximately 2\% higher $\overline{\Delta_{RGB}}_{,within}$ than their polynomial counterparts (FIN0 and FIN2).

TPS corrections achieved the best results, with TPS0 and TPS1 scoring less than 1\%. Despite earlier issues, TPS2 and TPS3, which approximated the TPS method to AFF3, also scored excellently, better than VAN3, the best polynomial correction. Increasing the smooth factor in the TPS formulation (TPS1 → TPS2 → TPS3) increased $\overline{\Delta_{RGB}}_{,within}$, smoothing the RGB space color deformation fitting.

On the other hand, results for \emph{inter-distances} $\overline{\Delta_{RGB}}_{,inter}$ (\autoref{fig:inter_distances_zoom}) showed that AFF corrections performed as expected, scoring similar $\overline{\Delta_{RGB}}_{,inter}$ to $\overline{\Delta_{RGB}}_{,within}$. The $\overline{\Delta_{RGB}}_{,inter}$ increased by around 1-2\% for all corrections compared to $\overline{\Delta_{RGB}}_{,within}$. AFF3 was the best, with a mean and median inner-distance around 4\%.

For polynomial corrections, VAN corrections presented similar results to AFF3, with mean $\overline{\Delta_{RGB}}_{,inter}$ around 4\% and matching distributions. However, VAN2 and VAN3 performed worse, with spread distributions and a mean $\overline{\Delta_{RGB}}_{,inter}$ around 8\%. Despite higher degree polynomial expansions reducing $\overline{\Delta_{RGB}}_{,within}$, they increased $\overline{\Delta_{RGB}}_{,inter}$. CHE corrections also matched AFF3's performance, with a mean and median inner-distance around 4\%. FIN corrections scored the worst; FIN0 and FIN1 were similar to AFF1 and AFF2, while FIN3 performed the worst overall, with a median $\overline{\Delta_{RGB}}_{,inter}$ of almost 10\% and a mean of nearly 8\%. Root-polynomial corrections presented worse results than their respective polynomial corrections.

TPS corrections achieved the best results for this metric. The impact of smoothing on TPS correction was minor, with all four corrections scoring a median and mean $\overline{\Delta_{RGB}}_{,inter}$ around 2\%.

All in all, TPS corrections proved to provide the best solution to color correct images in the dataset. The original~\cite{Menesatti2012} proposal (TPS0) worked slightly better than our first proposal of using the recommended RBF for 3D spaces (TPS1). Our Smooth-TPS3D proposals (TPS2 and TPS3) scored the subsequent best results for both metrics, $ \overline{\Delta_{RGB}}_{,within} $ and $ \overline{\Delta_{RGB}}_{,inter} $. VAN3 proved to be a good competitor in the within-distance metric, in contrast it had one of the poor results in the $ \overline{\Delta_{RGB}}_{,inter} $ metric. AFF3, VAN0, VAN1 and all CHE methods proved to be good competitors in the $ \overline{\Delta_{RGB}}_{,inter} $ metric, that is an interesting result as it opens the possibility to have fall-back methods if the TPS fails.

\subsection{Execution time performance}

\begin{figure*}[!ht]
    \centering
    \setlength{\belowcaptionskip}{-10pt}
    \includegraphics[width=\textwidth]{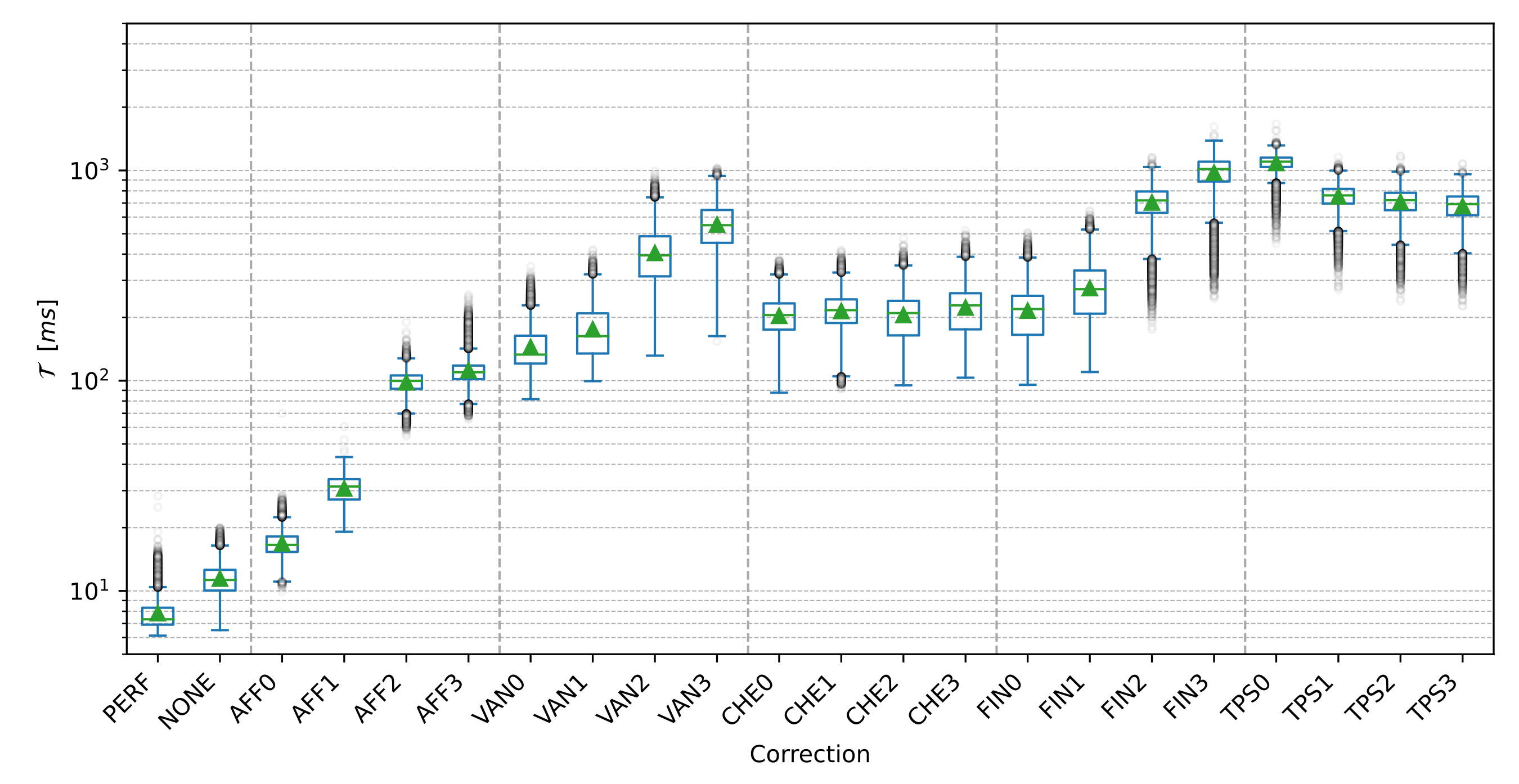}
    \caption{The execution time $ \mathcal{T} $ for each image in the dataset and is shown as a distribution against the color correction techniques. The distribution information is represented as follows: ($\triangle$) the mean, ($-$) the median at \textnum{Q2}, (box) from \textnum{Q1} to \textnum{Q3}, (whiskers) \textnum{1.5 \cdot (Q3 - Q1)} into \textnum{Q0} and \textnum{Q4} and ($\circ$) the outliers.}
    \label{fig:exec_time}
\end{figure*}

Regarding the \emph{execution time performance}, \autoref{fig:exec_time} shows the results of the measured execution times. PERF execution time represents the minimal time to compute our pipeline, as the PERF method also went all the way computing the same pipeline, it just returns the perfect expected image in 8-bit representation. NONE did the same but returning the image without applying any correction, note that this is slightly a slow process, as we are returning a different image for each augmentation.

Regarding affine corrections, all four corrections scored the best results in the benchmark, as expected, as they are the simpler corrections regarding implementation. AFF0 and AFF1 scored a mean $ \mathcal{T} $ per image around \textnum{20-30} ms. AFF2 and AFF3 scored around \textnum{100} ms.

For polynomial corrections, all four VAN corrections were slower than AFF3, with mean $ \mathcal{T} $ per image ranging from \textnum{100} ms to \textnum{600} ms. As AFF3 is a polynomial correction of order 1, and the subsequent corrections are VAN0 to VAN3, with degrees 2 to 5, respectively, we can observe that $ \mathcal{T} $ increases with the degree. This is also true for CHE and FIN corrections, which scored similar results to VAN1, with a mean $ \mathcal{T} $ per image around \textnum{200} ms. Superior limit is FIN3, that takes around  \textnum{1000} ms to compute an image. In FIN methods again, adding superior degrees to the polynomial expansion adds computational time to the correction. Also, adding complex operations to the pipeline, such as computing a root square, affects the computational cost of the solution.

Finally, for thin-plate corrections, TP0 scored very similar to FIN3, and in fact scored above the mean \textnum{1000} ms timestamp, achieving the worst score for this metric. Our proposal TPS1 reduced this computation time to around 800 ms. Also, adding smooth to the TPS method seems to reduce slightly its mean $ \mathcal{T}$, arriving down to \textnum{700} ms at TPS3. TPS3 is around \textnum{25} times slower than AFF1 and \textnum{8} times slower than AFF3.

All in all, results for AFF, VAN, CHE and FIN showed that increasing the degree of the polynomial expansion, increased the mean $ \mathcal{T} $ for each image. AFF corrections achieved the top scores as they are computationally simple. And, TPS scored poorly in this benchmark, as expected \cite{Luo2014}. Also, the scores for VAN2, VAN3, FIN2 and FIN3 were also poor. We accomplished to improve slightly the TPS computational performance by introducing a change in the RBF and the smooth parameter.

\section{Conclusions}

Here, we improved the work done by Menesatti et al.~\cite{Menesatti2012}. We successfully reproduced their findings about the TPS3D method for color correction to achieve image consistency in datasets. It can be shown that our results match theirs not only qualitatively but also quantitatively. For this purpose, \autoref{tab:resultssummary} shows a summary of the above-presented results for future comparison.

Also, we extended the study to other state-of-the-art methods, Gont et al.~\cite{Gong2013}, Chaung et al.~\cite{Cheung2004} and Finalyson et al.~\cite{Finlayson2015}. The TPS3D proved to be the best correction color method among the other in terms of color correction quality, both in $ \overline{\Delta_{RGB}}_{,within} $ and $ \overline{\Delta_{RGB}}_{,inter} $ metrics. Despite this, TPS3D is a heavy implementation compared to simpler methods such as AFF color corrections, resulting in $ \mathcal{T} $ per image \textnum{20} to \textnum{100} times higher.

Moreover, we proposed two criteria to detect failed corrections using the $ \overline{\Delta_{RGB}}_{,within} $ and $ \mathbf{\Delta_{RGB}}_{,pairwise} $ metrics. These criteria discovered failed corrections over the dataset that heavily affected TPS3D. Our proposal to approximate the TPS3D formulation by a smoothing factor proved the right way to systemically remove those ill-conditioned scenarios. And, we used different RBF in the TPS3D formulation. This did not mean a significant improvement in the color correction, although we did find that our proposed RBF would improve by a 30\% the results regarding the $ \mathcal{T} $ per image.

Finally, we demonstrated that $ \mathcal{T} $ increases linearly with the image size for all the compared color corrections, enabling to take into account this variable when designing future color correction pipelines.

Regarding future work to improve this color correction framework, \emph{systematic increase of color references should lead to a systematic improvement} this was not explored in the presented work. We could use novel colorimetric patterns, such as Pantone Color Match Card® or Color QR Codes~\cite{BenitoAltamirano2023}, as these patterns include several colors that are not present in the Macbeth ColorChecker. On the other hand, we could \emph{use different RBF} there exist several authors that have explored different RBF that could be placed in the kernel definition of the TPS3D method, here we only discussed between two RBF which were solutions for 2D and 3D for the thin-plate spline solution. Theoretically, any RBF could be used~\cite{Buhmann2000}, and even more modern smooth bump functions as well~\cite{Bita2015}.

\clearpage

\begin{table*}[h!t]
    \centering
    \caption{A summary of the presented results. The summary includes metrics for each color correction for 7 different metrics (see left), the within-distances and inter-distances also include some statistical information such as the mean ($ \upmu $), the standard deviation ($ \upsigma $) and the median ($\tilde{\upmu} $). The median should be considered as the reference figure in those metrics as their distributions are quite asymmetric.}
    \renewcommand{\arraystretch}{1.6}
    \scalebox{0.8}{
\begin{tabular}{l|r@{\hspace{1.2ex}}|r@{\hspace{1.2ex}}|rrr@{\hspace{1.2ex}}|r@{\hspace{1.2ex}}|rrr@{\hspace{1.2ex}}|r@{\hspace{1.2ex}}|r}
\hline 
criterion & $ \begin{array}{l}
     \overline{\Delta_{RGB}}_{,within} - \\
     \overline{\Delta_{RGB}}_{,within,NONE} \\ > 0 
\end{array} $ & 
$ \begin{array}{l}
     \min \left( \mathbf{\Delta_{RGB}}_{,pairwise} \right)    \\
     < \sqrt{3}
\end{array}  $ &  \multicolumn{4}{c|}{$ \overline{\Delta_{RGB}}_{,within} $} & 
\multicolumn{4}{c|}{$ \overline{\Delta_{RGB}}_{,inter} $}  & $ \mathcal{T} $      \\
\hline 
units & u. & - & - &  - & - & \% & - & - & - & \% & ms     \\
stats. & - & - & $ \upmu $ & $ \upsigma $ & $ \tilde{\upmu} $ &  $ \upmu $ &  $ \upmu $ & $ \upsigma $ & $ \tilde{\upmu} $ &  $ \upmu $ &  $ \upmu $   \\
\hline 
PERF &      0 &               0 &    0.99 &   0.12 &      0.99 &   0.223 &   0.945 &   0.019 &     0.949 &   0.214 &           8 \\
NONE &      0 &               0 &      59 &     27 &        56 &      13 &      52 &      23 &        50 &      12 &          11 \\
\hline
AFF0 &   5519 &            2018 &      37 &     21 &        36 &       8 &      39 &      22 &        37 &       9 &          17 \\
AFF1 &   2605 &            2152 &      24 &     18 &        21 &       5 &      24 &      19 &        22 &       6 &          30 \\
AFF2 &     38 &              24 &      22 &     14 &        21 &     5.1 &      30 &      17 &        28 &       7 &          96 \\
AFF3 &     19 &               9 &      12 &     10 &        10 &     2.7 &      17 &      12 &        15 &     3.9 &         110 \\
\hline
VAN0 &      9 &               1 &      10 &      8 &         8 &     2.2 &      20 &      22 &        14 &       4 &         142 \\
VAN1 &     10 &               2 &       8 &      7 &         6 &     1.9 &      20 &      23 &        14 &       5 &         172 \\
VAN2 &     35 &              23 &       8 &      7 &         5 &     1.8 &      30 &      40 &        20 &       8 &         397 \\
VAN3 &    176 &             154 &       8 &      8 &         5 &     1.7 &      40 &      40 &        20 &       9 &         542 \\
\hline
CHE0 &     10 &               2 &      11 &      9 &         9 &     2.6 &      17 &      15 &        14 &     3.9 &         199 \\
CHE1 &      9 &               1 &      10 &      9 &         8 &     2.3 &      18 &      18 &        14 &       4 &         210 \\
CHE2 &     11 &               2 &      10 &      8 &         8 &     2.2 &      18 &      22 &        13 &       4 &         202 \\
CHE3 &     16 &               6 &       9 &      7 &         7 &     2.0 &      20 &      25 &        14 &       5 &         219 \\
\hline
FIN0 &     56 &              46 &      14 &     11 &        13 &     3.2 &      26 &      24 &        21 &       6 &         212 \\
FIN1 &     29 &              21 &      19 &     12 &        18 &     4.4 &      29 &      18 &        27 &       7 &         271 \\
FIN2 &    537 &             462 &      10 &     11 &         6 &     2.2 &      40 &      40 &        20 &       8 &         691 \\
FIN3 &    346 &             193 &      17 &     11 &        15 &     3.9 &      42 &      34 &        34 &      10 &         958 \\
\hline
TPS0 &   8133 &            8117 &       2 &      7 &         2 &     0.5 &      10 &      13 &         7 &     2.3 &        1067 \\
TPS1 &   6359 &            6331 &       4 &     10 &         2 &     0.9 &      13 &      16 &         8 &       3 &         738 \\
TPS2 &      9 &               1 &     3.5 &    3.3 &       2.2 &     0.8 &      13 &      11 &        10 &     2.9 &         697 \\
TPS3 &     10 &               1 &       6 &      5 &         4 &     1.3 &      13 &      11 &        11 &     3.0 &         665 \\
\hline 

\end{tabular}
    }
    \label{tab:resultssummary}
\end{table*}


\section*{Acknowledgments}
This work has been funded in part by the European Research Council under the H2020 Framework Program ERC Grant Agreements no.727297 and no. 957527. I. Benito-Altamirano and C. Ventura acknowledge the support from projects PID2022-138721NB-I00, funded by MCIN (Spain) and PLEC2021-007868, funded by MCIN and by the EU. J.D. Prades  acknowledges the support from the sponsorship of the Alexander von Humboldt Professorship of the Humboldt Foundation and the Federal Ministry for Education and Research (Germany). C. Fàbrega acknowledges the support from Agència per la Competitivitat de l’Empresa (ACCIO, Innotec 2021 ACE034/21/000057).


%
%
\bibliographystyle{splncs04}
\bibliography{main}
\end{document}